\documentclass[runningheads]{llncs}
\usepackage{accv}
\usepackage{accvabbrv}
\usepackage{graphicx}
\usepackage{booktabs}
\usepackage{array}
\usepackage[accsupp]{axessibility}
\usepackage{hyperref}
\usepackage{orcidlink}

\newcommand{\cls}[1]{\textsc{#1}}

\begin{document}

\title{Different Timescales, Different Signals: Multimodal Landmark-Visibility Prediction in a Long-Term Webcam Network}
\titlerunning{Different Timescales, Different Signals}

\author{Bryceton Bible\inst{1} \and Hairong Qi\inst{1} \and Masahiro Toyoura\inst{2}}
\authorrunning{B. Bible et al.}
\institute{University of Tennessee, Knoxville, Knoxville, TN, USA
\email{bbible3@vols.utk.edu, hqi@utk.edu}
\and University of Yamanashi, Kofu, Japan
\email{mtoyoura@yamanashi.ac.jp}}

\maketitle

\begin{abstract}
Scenic visibility is not a single forecasting problem: planning a trip months ahead, deciding whether to go outside today, and judging whether tomorrow will be clear rely on different information. Using a year-scale multimodal archive of Mount Fuji webcams and meteorological observations, we identify a clear hierarchy across timescales. Seasonal climatology provides useful long-range guidance, webcam imagery is the strongest same-day observation (AUC 0.95), and the numerical forecast valid on the target day carries most of the day-specific signal one to three days ahead. Revealing this hierarchy requires evaluation that respects the dependence among cameras observing the same regional weather state: permissive splits substantially overstate multi-day performance, while date-grouped and forward-in-time evaluation reduce the apparent contribution of the current scene and current weather. On 6{,}575 camera-days, the current visual summary provides no measurable one-day gain once the target-day forecast is available. The remaining opportunity lies at shorter horizons, where the present visual representation discards cloud geometry and motion. These results recast scenic visibility prediction as a multi-timescale sensing problem rather than a single multimodal forecasting task.
\keywords{multimodal learning \and environmental monitoring \and webcams \and weather forecasting \and multi-timescale prediction \and temporal evaluation}
\end{abstract}

\section{Introduction}
\label{sec:intro}

A scenic-visibility forecast can mean very different things at different horizons. A traveler planning a visit months ahead needs a seasonal prior; a person deciding whether to go outside now can look directly at the mountain; and someone choosing whether tomorrow morning is promising must rely on a forecast of future atmospheric conditions. More broadly, environmental monitoring systems must distinguish between observations that characterize current conditions and those that provide predictive information at longer horizons. The utility of each sensing modality depends not only on what it measures, but also on the temporal scale of the prediction task.

Fixed cameras are well suited to studying this distinction. Unlike point instruments, they continuously and cheaply record the appearance of a whole scene as the atmosphere between the lens and a distant target changes. Meteorological visibility has been estimated from cameras for more than a decade~\cite{babari2012visibility,graves2014camera,palvanov2019visnet}, and fixed outdoor cameras and large image collections have supported studies of scene appearance and weather across sites and time~\cite{jacobs2007consistent,islam2013images,glasner2015hot,chu2017camera}. Mount Fuji is an especially useful target: it is visible from a hundred kilometres away on the right day, watched by dozens of public webcams, and has a decades-long human record of viewing days used as a proxy for regional air quality~\cite{kurashige1998howmany}.

The original FujiView study introduced scenic visibility forecasting as a multimodal learning task~\cite{bible2026fujiview}, using several months of webcam and weather observations to show that late fusion could predict whether the mountain would be visible today and on following days. A year-scale archive now lets us ask a broader longitudinal question: \emph{which signals remain useful at which timescale?} The longer record exposes strong seasonality, repeated views of the same regional weather state, real sensor gaps, and enough temporal span to test future-facing generalization. Across these conditions, a simple hierarchy emerges: climatology for long-range planning, the camera for the present day, and numerical forecasts for future days.

Establishing that hierarchy requires more than rerunning the original model on more data. We evaluate 6{,}575 camera-days under six schemes ranging from permissive folds to an expanding window trained strictly on the past, separate seasonal discrimination from within-month day ranking, and align archived forecast fields to their actual valid dates. These controls show why some earlier-looking gains do not survive stricter evaluation, but the correction is a means to the larger result: the useful modality changes with the decision horizon.

This paper makes three contributions. First, it provides a validated year-scale camera-day benchmark with paired imagery, meteorology, human labels and real missing-data patterns (\cref{sec:dataset}). Second, it identifies a multi-timescale structure in scenic visibility prediction: climatology supports long-range planning, webcams dominate same-day nowcasting, and target-day numerical forecasts carry most of the future day-specific signal (\cref{sec:results}). Third, it shows how date dependence, seasonality, forecast valid-time alignment and forward-time evaluation must be handled to reveal that structure reliably, while identifying local cloud state and motion as the missing visual information for the next, sub-day forecasting regime (\cref{sec:method,sec:discussion}).

\section{Related Work}
\label{sec:related}

\paragraph{Cameras as weather and visibility sensors.}
Atmospheric visibility has been estimated from vehicle-mounted and fixed ordinary cameras with contrast-based physical models~\cite{hautiere2006automatic,babari2011model} and with estimators calibrated or learned against reference measurements, from regression on contrast features to deep classifiers~\cite{babari2012visibility,graves2011visibility,graves2014camera,li2017meteorological,palvanov2019visnet}. Other methods learn visibility from relative image comparisons rather than requiring an absolute numeric label for every image~\cite{lepetit2020learning,you2019relative}. Recent work has explored multimodal visibility estimation using complementary image modalities~\cite{wang2020multimodal} and combinations of imagery with meteorological variables~\cite{kopecka2026estimation}. Landmark-based methods restrict a learned estimator to static landmark regions of a fixed camera view~\cite{lo2025meteorological}. Multi-site webcam archives document weather- and season-related appearance variation~\cite{jacobs2007consistent,islam2013images,laffont2014transient}, and phenology and camera-trap networks treat cameras as scientific infrastructure~\cite{richardson2018tracking,seyednasrollah2019phenocam,swanson2015snapshot,norouzzadeh2018automatically}. Here the quantity is the perceived visibility of one landmark from many viewpoints, observed long enough to be forecast and verified.

\paragraph{Fusion, post-processing and missing modalities.}
In the early-versus-late fusion distinction~\cite{snoek2005early} and the taxonomies of multimodal learning~\cite{baltrusaitis2019multimodal,liang2024foundations}, our design is a model-agnostic, late-fusion-style combination of a vision expert's predictions with tabular weather features; joint training of heterogeneous modalities is known to be difficult~\cite{wang2020what}, and boosted trees~\cite{ke2017lightgbm} remain competitive on tabular data of comparable size~\cite{grinsztajn2022why,shwartzziv2022tabular}. Statistical post-processing of numerical forecasts has a long tradition~\cite{glahn1972mos} and is now often learned~\cite{rasp2018neural,gronquist2021deep}, in line with the call to combine data-driven models with physical forecasts~\cite{reichstein2019deep}; tree models have also been used to post-process numerical weather-prediction outputs for short-term visibility prediction~\cite{kim2022shortterm}. Sensors of a real network fail, and multimodal models have been made robust to missing inputs, for example by dropping modalities during training~\cite{neverova2016moddrop} or by learning to reconstruct the missing features~\cite{ma2021smil}.

\paragraph{Evaluation under dependence.}
Observations that share a site, a date or a season are not independent, and cross-validation that ignores the dependence overestimates skill~\cite{roberts2017crossvalidation,meyer2018improving}; leakage of this kind is a documented cause of irreproducible results in machine-learning-based science~\cite{kapoor2023leakage}. Benchmarks formalise shifts across sites and time~\cite{beery2018terra,koh2021wilds,yao2022wildtime}, forecast evaluation distinguishes several notions of forecast quality and value~\cite{murphy1993good}, and impurity-based feature importance is known to mislead~\cite{strobl2007bias}. We apply these lessons to a multi-camera, multimodal setting in which they interact.

\section{The FujiView Archive as a Multimodal Dataset}
\label{sec:dataset}

\begin{figure}[tp]
\centering
\includegraphics[width=\textwidth]{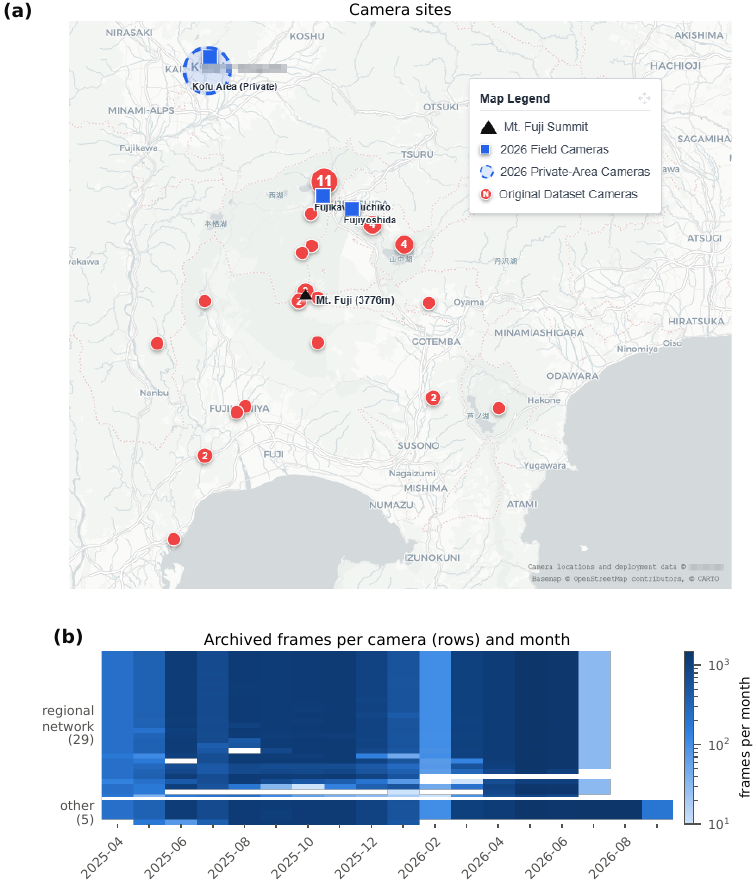}
\caption{The FujiView archive. (a) Approximate positions of the public webcams configured in the collector, on an OpenStreetMap/CARTO basemap: the 41 sources with place-level coordinates (red; nearby cameras are clustered and the marker carries their number; one source east of the displayed region is omitted). Not every configured source produced archived frames; panel (b) shows the 34 that did. The blue markers are cameras installed in 2026 for a separate study; they are not part of the archive analysed here. (b) Archived frames per camera and month; the sparse July 2026 column and the empty columns after it reflect an upstream format change that interrupted most feeds until a repair in September 2026, after the listing shown.}
\label{fig:dataset}
\end{figure}

\begin{table}[t]
\centering
\caption{Components of the FujiView archive used in this paper. A weather row holds, for one camera and one collection cycle, the current conditions and six daily forecast blocks; the total is an estimate (file count times rows per file for the part held in the cloud).}
\label{tab:dataset}
\scriptsize
\setlength{\tabcolsep}{3pt}
\begin{tabular}{@{}>{\raggedright\arraybackslash}p{2.2cm}>{\raggedright\arraybackslash}p{3.2cm}>{\raggedright\arraybackslash}p{3.25cm}>{\raggedright\arraybackslash}p{2.8cm}@{}}
\toprule
Component & Source & Size & Period \\
\midrule
Webcam frames & 34 cameras, 30-min cadence & 424,853 frames (197 GB) & Apr 2025 -- Sep 2026 \\
\addlinespace[2pt]
Weather rows & Open-Meteo, per camera location and cycle & $\approx$708,818 rows (current conditions and six daily forecast blocks) & May 2025 -- Sep 2026 \\
\addlinespace[2pt]
Human labels & 22 labelers, five ordinal classes & 23,894 labels on 22,884 frames & Apr 2025 -- Feb 2026 (capture of the frames) \\
\addlinespace[2pt]
Scored frames (this paper) & frozen visual classifier, daytime frames & 141,816 readable of 142,115 & May 2025 -- Apr 2026 \\
\addlinespace[2pt]
Camera-day table (this paper) & frame scores joined with the 09:00 weather row & 6,575 camera-days, 32 cameras, 231 dates, 15 weather locations & May 2025 -- Apr 2026 \\
\bottomrule
\end{tabular}

\end{table}

\paragraph{Frames.}
A scheduled collector fetches one frame from every configured public webcam around Mount Fuji every 30 minutes. As of September 2026, the archive contains 424{,}853 frames (197~GB) captured between 25 April 2025 and 5 September 2026; 34 of 43 configured sources produced archived frames (\cref{fig:dataset}). Most cameras belong to one regional live-camera network; four are municipal or commercial cameras to the west, south and east of the mountain, and one is a video stream. Frames are typically $3840\times2160$ pixels; the summit fills them as a large central cone or as a few hundred pixels above a ridge.

\paragraph{Weather.}
With every collection cycle the collector queries the Open-Meteo forecast API~\cite{zippenfenig2024openmeteo} at the place-level coordinates configured for each camera and stores one row per camera: the current conditions of the provider's model (temperature, apparent temperature, humidity, cloud cover, precipitation, rain, snowfall, sea-level and surface pressure, wind speed and gusts, a daylight flag and a weather code) and six blocks labeled as daily forecasts at the clock hour of the fetch. The 41 configured entries share 20 distinct coordinates, so the cameras of one lake shore receive identical rows. The local mirror used here holds 406{,}935 rows from 27 May 2025 to 10 April 2026; the cloud collector that replaced it has written about 302{,}000 more (\cref{tab:dataset}). Every row was fetched before the time at which we let a model decide.

\begin{figure}[t]
\centering
\includegraphics[width=0.76\textwidth]{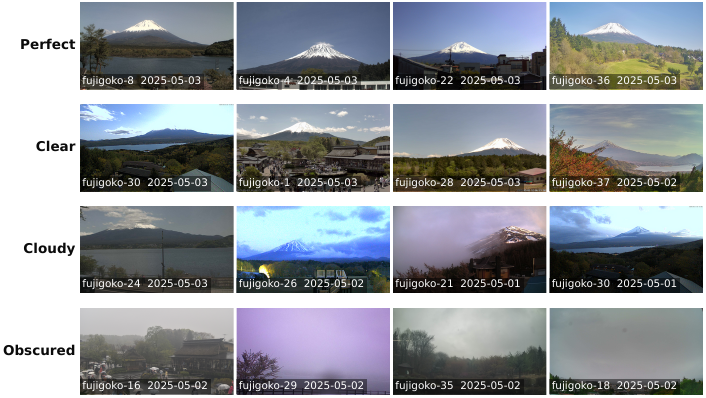}
\caption{The four visibility classes on frames of the regional camera network, four viewpoints per class. Each frame carries a reference label that at least two further labelers confirmed.}
\label{fig:examples}
\end{figure}

\paragraph{Human labels.}
Human annotations use five ordered categories: \cls{Perfect} (summit fully visible, no cloud overlap), \cls{Clear} (visible, minimal overlap), \cls{Cloudy} (visible, substantial overlap), \cls{Obscured} (mostly or completely hidden), and \cls{Bad} for unusable frames (\cref{fig:examples}). An annotation campaign with 22 labelers produced 23{,}894 labels on 22{,}884 frames captured between April 2025 and February 2026. The labels serve two purposes here: a four-class subset of 14{,}049 frames was used to train and validate the visual classifier, and the labels of daytime frames give the human day labels against which the model-derived targets are validated. For the downstream day-level target, we therefore use only the coarse grouping of \cls{Perfect} and \cls{Clear} as \emph{visible}, rather than relying on distinctions among all four ordered visibility classes.

\paragraph{Gaps.}
Like any long-running sensor deployment, the archive contains gaps. Collector outages removed whole weeks (July and December 2025, mid-January to early March 2026), and individual cameras fail for days or months. On the dates on which the collector ran during the period analysed here, 17\% of the possible camera-days have a weather row but no usable camera day, 0.3\% the reverse, and 12\% neither.

\section{Method}
\label{sec:method}

\begin{figure}[t]
\centering
\includegraphics[width=0.92\textwidth]{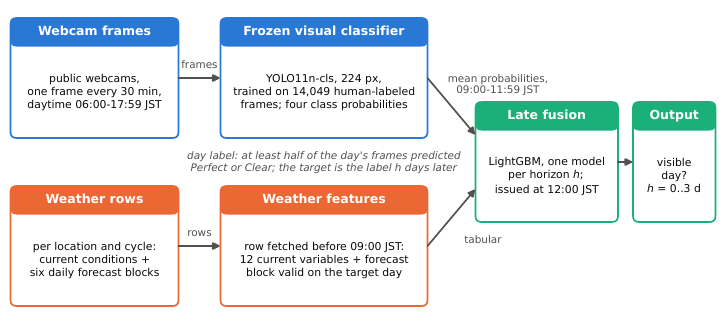}
\caption{The multimodal pipeline as evaluated. A frozen visual classifier scores every daytime frame; the mean class probabilities of the 09:00--11:59 JST window, the weather row fetched before 09:00 and the forecast block valid on the target day feed one gradient-boosted classifier per horizon.}
\label{fig:pipeline}
\end{figure}

\paragraph{Visual model and day labels.}
The visual component is a YOLO11n classification network~\cite{jocher2024yolo11} with ImageNet initialisation, trained and validated at $224\times224$ input on 14{,}049 labeled frames and frozen for everything that follows. On the 2{,}884 daytime frames of a held-out set, formed from whole seven-day blocks that contributed no frame to training, it reproduces the human label in 72.8\% of the cases and the coarse visible/not-visible decision in 91.7\%; we report this only because the day labels are derived from the classifier. We score every archived frame of the local mirror captured between 06:00 and 17:59 JST from 27 May 2025 to 10 April 2026, 141{,}816 readable frames from 34 cameras (\cref{fig:pipeline}). Following the original study, a camera-day is \emph{visible} when at least half of its scored frames are predicted \cls{Perfect} or \cls{Clear}; in addition, we drop days with fewer than six scored frames. One lakeside camera whose field of view does not contain the summit is excluded from every analysis, because its labels can only describe the weather, and one camera has too few frames for a single day. The target of horizon $h\in\{0,1,2,3\}$ is the label of the same camera $h$ calendar days after the decision day.

\paragraph{Feature blocks.}
A prediction is issued at noon from three possible inputs. The \emph{visual} block summarises the camera frames from 09:00--11:59 JST, the \emph{current-weather} block uses the most recent weather row available before 09:00, and the \emph{forecast} block uses the stored numerical forecast for the target day. We verified the forecast valid times from both the collector logic and the archived observations. A stored block labeled as $k$ days ahead is actually valid $k-1$ days ahead; for example, the first block closely matches same-day conditions rather than the following day. We therefore pair horizon $h$ with block $h+1$ and retain the historical pairing as a sensitivity comparison. The longest-horizon forecast fields also differ systematically from the shorter-horizon fields, so we treat the three-day result cautiously.

\paragraph{Learner.}
One LightGBM classifier~\cite{ke2017lightgbm} is fitted per horizon and feature set with fixed hyper-parameters (learning rate 0.05, 63 leaves, feature and bagging fractions 0.9, at least 40 samples per leaf, balanced class weights) and early stopping on an inner 15\% of the training rows. The deliberately simple, untuned learner compares information sources rather than introducing a new fusion architecture; fixing its configuration avoids confounding modality and split comparisons with model selection. Every main fit and block ablation is repeated with five seeds; the explicitly labeled sensitivity designs and human-label checks use a single seed.

\paragraph{Evaluation schemes.}
Six schemes differ in what training and test are allowed to share. \emph{Random days} uses five shuffled folds over camera-days as a permissive reference. \emph{Unseen cameras}, \emph{unseen dates}, \emph{unseen weeks} and \emph{unseen months} group the folds by the corresponding unit. These grouped schemes prevent some forms of overlap, but they can still train on dates that follow their test dates. \emph{Chronological} evaluation instead uses monthly test blocks from August 2025 to April 2026 (February 2026 has no usable camera-day and therefore no block), training only on earlier camera-days whose targets also precede the test block. Early stopping always uses training rows only.

\paragraph{References and metrics.}
Three references need no fitted model of the features: the majority class of the training rows; persistence, the label of the last completed day before the decision; and the cloud cover of the forecast block (of the current block at $h=0$) used directly as a score. We report accuracy at a threshold of 0.5 and ROC-AUC as the mean over folds or blocks and seeds. Because a fold of the first five schemes mixes the seasons, we also compute AUC within each calendar month of the decision date from the test predictions of a scheme and take an unweighted mean over the eight months used by the chronological design. For grouped schemes, predictions from the relevant fold models are pooled within each month before scoring. This \emph{within-month AUC} asks whether days are ranked correctly among other days in the same month; it suppresses the easiest annual-cycle discrimination but does not remove every within-month seasonal trend. Intervals are 95\% percentile intervals of a bootstrap that resamples whole decision dates within months (2{,}000 resamples); differences between feature sets are bootstrapped in pairs.

\section{Results}
\label{sec:results}

\begin{table}[t]
\centering
\caption{Within-month ROC-AUC for three representative evaluation schemes, making the columns directly comparable across split designs. Random days permits different cameras from the same date to appear across folds; unseen dates holds out whole calendar dates; chronological trains only on earlier data. Weather combines current weather with the target-day forecast for $h>0$ and uses current weather at $h=0$; fusion adds the visual block. The full six-scheme comparison, including overall fold/block AUC, is shown in \cref{fig:skill}.}
\label{tab:main}
\scriptsize
\setlength{\tabcolsep}{10pt}
\begin{tabular}{@{}llrrr@{}}
\toprule
Evaluation & $h$ & Visual & Weather & Fusion \\
\midrule
random days & 0 & 0.95 & 0.93 & 0.97 \\
 & 1 & 0.55 & 0.92 & 0.92 \\
 & 2 & 0.56 & 0.92 & 0.93 \\
 & 3 & 0.56 & 0.93 & 0.93 \\
\midrule
unseen dates & 0 & 0.94 & 0.81 & 0.94 \\
 & 1 & 0.52 & 0.76 & 0.76 \\
 & 2 & 0.52 & 0.75 & 0.76 \\
 & 3 & 0.53 & 0.71 & 0.70 \\
\midrule
chronological & 0 & 0.95 & 0.79 & 0.94 \\
 & 1 & 0.51 & 0.68 & 0.69 \\
 & 2 & 0.51 & 0.75 & 0.76 \\
 & 3 & 0.53 & 0.65 & 0.66 \\
\bottomrule
\end{tabular}

\end{table}

\begin{figure}[t]
\centering
\includegraphics[width=\textwidth]{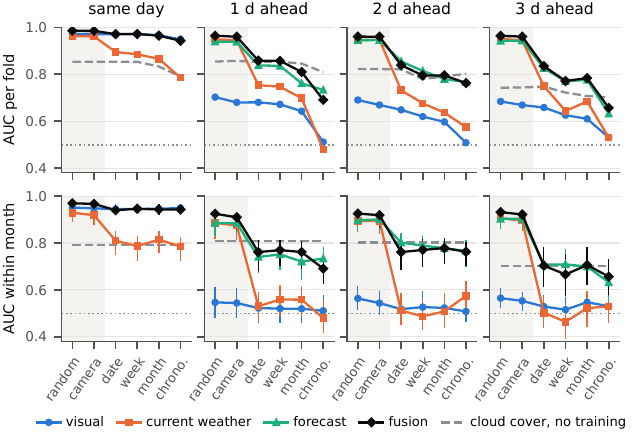}
\caption{ROC-AUC of the feature blocks under the six schemes, ordered from the most permissive to the strictest. Top: mean over folds or blocks. Bottom: within calendar months, with 95\% bootstrap intervals. Shaded: schemes in which training and test share calendar dates. Dashed: cloud cover of the forecast (of the current conditions on the same day) used as a score without training.}
\label{fig:skill}
\end{figure}

\paragraph{A year-scale camera-day benchmark.}
The evaluation table holds 6{,}575 camera-days from 32 cameras and 231 dates between 28 May 2025 and 10 April 2026. Visibility varies strongly with season, from infrequent summer views to much clearer winter conditions. On camera-days with enough human annotations, the model-derived day label agrees with the human day label on 89.7\% of cases, and agreement remains essentially unchanged on weeks withheld from classifier training. The model-derived targets are therefore a slightly conservative but useful proxy for human judgment at the day level.

\paragraph{The archive contains multiple predictive timescales.}
Cameras observing the same calendar day usually agree, reflecting a regional atmospheric state shared across viewpoints. Season also explains a substantial part of visibility: calendar month alone accounts for about a quarter of the variation in the camera-day label. Apparent day-to-day persistence is much weaker once monthly means are removed. These patterns already suggest that long-range planning, same-day observation and future-day forecasting should not be treated as the same prediction problem.

\paragraph{Today: the camera is the strongest observation.}
For same-day visibility, the visual block is consistently the strongest input, reaching 0.95 AUC under strict chronological evaluation (\cref{tab:main,fig:skill}). Current weather does not measurably improve on the camera within a month, and the visual result is stable across cameras and when scored against human day labels. Because the whole-day target includes the morning frames used by the visual summary, this is best interpreted as a nowcast rather than a forecast. Rebuilding the target from afternoon-only frames reduces performance, but the camera remains strongly informative.

\paragraph{Why permissive splits blur the timescale hierarchy.}
One to three days ahead, random folds make the weather features look extremely strong, with overall AUC around 0.96 (\cref{fig:skill}). Grouping by camera barely changes this because every calendar date is still represented in training. The cameras share the same regional weather and, in several cases, identical weather rows; a model can therefore recognize conditions from a date it has effectively already seen. Holding out whole dates lowers performance substantially, and the within-month one-day AUC falls from 0.92 to 0.76. The problem is the split design, not one particular weather variable.

\paragraph{Long-range planning: seasonality is useful, but it is a different task.}
Seasonality is genuine predictive information: Fuji is much more often visible in winter than in summer, and calendar month alone explains about a quarter of the variation in the target. That is useful for the question \emph{when in the year should I plan a trip?}, but it does not answer \emph{will this particular day be visible?}. Date-, week- and month-blocked folds still mix seasons across training and test, so seasonal cues can inflate overall AUC even when they do little to rank nearby days. Within-month evaluation removes most of this advantage: current weather and the current camera view fall close to chance for future-day prediction (\cref{tab:blocks}).

\paragraph{Long-range planning: the visibility climatology.}
\begin{figure}[t]
\centering
\includegraphics[width=\textwidth]{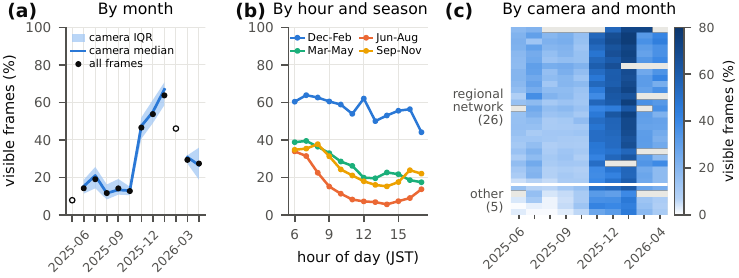}
\caption{Share of daytime frames predicted \cls{Perfect} or \cls{Clear} by the frozen visual model. (a) By month: all frames (open markers: months with incomplete coverage) and the median and interquartile range over cameras. (b) By hour of day and season. (c) By camera (rows) and month, for cells with at least 100 frames.}
\label{fig:climatology}
\end{figure}

Scoring every daytime frame also yields a description of the observable itself. Across the 137{,}916 scored daytime frames that remain after the lakeside camera is excluded, 26.9\% show the summit as \cls{Perfect} or \cls{Clear}. The seasonal cycle is pronounced (\cref{fig:climatology}a): visibility is lowest through summer and early autumn and rises sharply in winter, reaching 64\% in January. A quantitative comparison with the decades-long human record of Fuji-viewing days from the Tokyo area~\cite{kurashige1998howmany} is left open because viewpoint and definition differ. The diurnal pattern also changes with season (\cref{fig:climatology}b): summer visibility is best early in the morning and falls sharply later in the day, whereas winter remains comparatively clear across daylight hours. Cameras broadly agree on these patterns but differ in their absolute visibility rates (\cref{fig:climatology}c). These are model predictions rather than human observations, but they provide continuous coverage across cameras and daylight hours.

\begin{table}[t]
\centering
\caption{Within-month ROC-AUC for the core future-day feature sets under unseen-date (D) and chronological (C) evaluation. ``Cloud cover, no training'' uses $1-\mathrm{cloud\ cover}/100$ directly. The historical pairing uses a forecast valid one day before the target date.}
\label{tab:blocks}
\scriptsize
\setlength{\tabcolsep}{6pt}
\begin{tabular}{@{}lrrrrrr@{}}
\toprule
 & \multicolumn{2}{c}{1 d ahead} & \multicolumn{2}{c}{2 d ahead} & \multicolumn{2}{c}{3 d ahead} \\
\cmidrule(lr){2-3}\cmidrule(lr){4-5}\cmidrule(l){6-7}
Features & D & C & D & C & D & C \\
\midrule
Cloud cover, no training & 0.81 & 0.81 & 0.80 & 0.80 & 0.70 & 0.70 \\
Forecast block & 0.74 & 0.73 & 0.80 & 0.77 & 0.71 & 0.63 \\
Current weather & 0.53 & 0.48 & 0.51 & 0.58 & 0.50 & 0.53 \\
Visual & 0.52 & 0.51 & 0.52 & 0.51 & 0.53 & 0.53 \\
Fusion & 0.76 & 0.69 & 0.76 & 0.76 & 0.70 & 0.66 \\
Fusion, historical pairing & 0.54 & 0.50 & 0.49 & 0.57 & 0.50 & 0.53 \\
\bottomrule
\end{tabular}

\end{table}

\paragraph{Future days: the target-day forecast carries the signal.}
For future days, the target-day numerical forecast is the only input with consistently strong within-month skill across evaluation schemes (\cref{tab:blocks}). Adding it to current weather improves one-day within-month AUC by about 0.2, while adding the current camera view produces no measurable gain. A particularly strong reference is forecast cloud cover used directly, without any learned model: it reaches 0.81 within-month AUC one day ahead. Under chronological evaluation, the full fusion model does not beat that simple reference and is lower by 0.12 at one day. The result is therefore not that multimodal fusion is universally better, but that the useful modality depends on the horizon.

\paragraph{Historical feature alignment as a sensitivity check.}
The historical feature pairing selects a forecast valid one day before the target date and, at one day ahead, effectively reuses the current conditions (\cref{sec:method}). Under unseen-date and chronological evaluation this pairing shows little within-month future-day skill, even though permissive folds still score highly. Aligning the forecast to the actual target date improves one-day within-month AUC by about 0.2. The original study's same-day visual result is reproduced at scale~\cite{bible2026fujiview}. Because the published split description does not map exactly onto the schemes used here, we treat the multi-day comparison as a sensitivity analysis rather than an exact rerun.

\begin{table}[t]
\centering
\caption{Missing-modality accuracy under unseen-date evaluation at three stress levels. Camera missing is evaluated on the same day; weather missing is evaluated one day ahead. Naive applies the fused model to incomplete input; spec. falls back to the remaining-modality specialist; drop. uses a fused model trained with modality dropout.}
\label{tab:missing}
\scriptsize
\setlength{\tabcolsep}{7pt}
\begin{tabular}{@{}rrrrrrr@{}}
\toprule
 & \multicolumn{3}{c}{Camera missing, same day} & \multicolumn{3}{c}{Weather missing, 1 d ahead} \\
\cmidrule(lr){2-4}\cmidrule(l){5-7}
Missing & naive & spec. & drop. & naive & spec. & drop. \\
\midrule
0\% & 0.93 & 0.93 & 0.93 & 0.84 & 0.84 & 0.84 \\
50\% & 0.86 & 0.89 & 0.89 & 0.54 & 0.76 & 0.77 \\
100\% & 0.80 & 0.85 & 0.85 & 0.23 & 0.68 & 0.69 \\
\bottomrule
\end{tabular}

\end{table}

\paragraph{Missing modalities.}
The common failure of the network is a camera without a usable day while the weather row exists (\cref{sec:dataset}). We remove the visual block or the whole weather input from a share of the test camera-days and compare three answers (\cref{tab:missing}): the fused model on the incomplete input, a fallback to the specialist of the remaining modality, and a fused model trained on rows from which one modality had been removed, in the spirit of modality dropout~\cite{neverova2016moddrop}. The naive answer fails badly: the fused model never saw a missing value in training, the learner then treats a missing number as zero, and with the weather input missing the accuracy one day ahead falls from 0.84 to 0.23. The specialist fallback and the dropout-trained model both degrade to the level of the remaining modality and no further, and dropout costs nothing when the input is complete. Because only one block matters at a horizon, a missing camera costs nothing beyond the same day, and a missing weather row costs little on the same day (0.92 accuracy).

\paragraph{Robustness.}
Results are stable across random seeds and alternative chronological designs: the same one-day ordering persists, with current weather and the camera near chance within months, the target-day forecast clearly stronger, and fusion in between. Changing the inner early-stopping split has only a small effect. Extending the forecast block with pressure and wind does not change the qualitative ranking, although it modestly improves the three-day result under unseen-date evaluation.

\section{Discussion}
\label{sec:discussion}

\paragraph{A hierarchy of forecasting tasks.}
The longer archive shows that visibility prediction in this network should not be treated as one homogeneous zero-to-three-day problem. If the question is \emph{when in the year should I plan a Fuji-viewing trip?}, season and month already provide a strong climatological prior. If the question is \emph{can I see Fuji today?}, the webcam is the most useful direct observation. If the question is \emph{will Fuji be visible tomorrow?}, the numerical forecast for tomorrow carries most of the day-specific signal. Multimodality can therefore be valuable at the system level without guaranteeing gains from fusing every feature: here imagery is strong for nowcasting but its static class-probability summary adds no measurable one-day skill once the target-day forecast is provided.

\paragraph{What is missing at shorter horizons.}
The hierarchy also reveals the question that the current system does not yet answer well: \emph{will the clouds visible now cover Fuji in the next few hours?} The present visual block compresses a morning sequence into visibility-class probabilities and discards cloud type, geometry, direction and motion. Those properties are physically plausible predictors at horizons such as 30 minutes to six hours. Future work could investigate richer spatial and temporal representations, including cloud-motion features from webcam sequences and higher-dimensional meteorological inputs such as weather radar, to determine whether they provide complementary information beyond existing numerical forecasts.

\paragraph{Lessons for evaluating multi-sensor forecasts.}
Three practical lessons follow. First, when multiple sensors observe the same regional event, split by the event---in our case, calendar date---not only by sensor. Second, when the target has strong seasonality, evaluate discrimination within comparable periods so that easy seasonal differences do not dominate the score. Third, verify the valid time and standalone predictive strength of external forecast inputs before attributing their information to a learned fusion model. Although our empirical findings are specific to Mount Fuji, the evaluation principles---accounting for shared environmental conditions, temporal dependence, and the standalone predictive value of each modality---are applicable to other environmental monitoring problems. Replication across additional landmarks and sensor networks would help establish how the observed modality hierarchy changes with geography and local weather conditions.

\paragraph{Why the learner does not beat its input.}
The paired multimodal evaluation covers May 2025 through April 2026, so each season is represented at most once; a learned correction to an already informative numerical forecast may not generalize across months. The evidence does not isolate one cause: removing temperature and pressure changes performance only slightly, and even a model trained on forecast cloud cover alone underperforms the raw cloud-cover score. The same ordering appears against human day labels, arguing against label derivation as the sole explanation. More annual cycles and richer, motion-aware visual representations are needed to test when fusion adds information beyond the numerical forecast.

\section{Limitations}
\label{sec:limitations}

The paired multimodal evaluation covers May 2025 through April 2026 in one region, so each season is represented at most once and the chronological test set contains relatively few months. Data collection is ongoing, so future work can repeat the same evaluation across multiple annual cycles as additional paired observations accumulate. Day-level targets are derived from a visual classifier trained partly on labels from the same camera network, although we validate them against human labels and against weeks withheld from classifier training. The temporal split constrains the fusion learner but not the frozen visual classifier, and we evaluate only one visual backbone (YOLO11n) and its class-probability summary. Thus the lack of added day-ahead visual skill cannot be generalized to richer visual encoders or motion-aware features without further tests. Weather forecasts come from a single provider, and several cameras share forecast coordinates; the archived three-day-ahead fields also differ systematically from shorter-horizon fields, so we treat that horizon cautiously. Finally, the learner uses the original hyper-parameters without tuning, so our conclusions concern the information available to this formulation rather than every possible multimodal model. The within-month metric and training-free forecast baseline were added after the initial chronological analysis and should be viewed as exploratory controls.

\section{Conclusion}
\label{sec:conclusion}

Scenic visibility is best understood as a hierarchy of forecasting tasks rather than one homogeneous multimodal problem. In the year-scale FujiView archive, seasonal context supports long-range planning, the camera is a strong same-day nowcast, and the numerical forecast valid on the target day carries most day-specific future discrimination. Date dependence, seasonal structure and forecast valid-time alignment can obscure that division if they are not handled explicitly, so date-grouped, within-period and forward-time evaluation are necessary tools for measuring the hierarchy honestly. The resulting picture also identifies what the present formulation still lacks: a static visibility summary describes the mountain now but discards the local cloud state and motion most likely to matter at sub-day horizons. The next stage is therefore not simply more fusion, but matching each forecast horizon to the observations that can physically inform it. The camera-day table, predictions and analysis code will be released so that this longitudinal evaluation can be repeated as the archive grows.

\clearpage
\bibliographystyle{splncs04}
\bibliography{references}

\begin{thebibliography}{10}
\providecommand{\url}[1]{\texttt{#1}}
\providecommand{\urlprefix}{URL }
\providecommand{\doi}[1]{https://doi.org/#1}

\bibitem{babari2011model}
Babari, R., Hauti{\`e}re, N., Dumont, {\'E}., Br{\'e}mond, R., Paparoditis, N.: A model-driven approach to estimate atmospheric visibility with ordinary cameras. Atmospheric Environment  \textbf{45}(30),  5316--5324 (2011). \doi{10.1016/j.atmosenv.2011.06.053}

\bibitem{babari2012visibility}
Babari, R., Hauti{\`e}re, N., Dumont, {\'E}., Paparoditis, N., Misener, J.: Visibility monitoring using conventional roadside cameras -- emerging applications. Transportation Research Part C: Emerging Technologies  \textbf{22},  17--28 (2012). \doi{10.1016/j.trc.2011.11.012}

\bibitem{baltrusaitis2019multimodal}
Baltru{\v{s}}aitis, T., Ahuja, C., Morency, L.P.: Multimodal machine learning: A survey and taxonomy. IEEE Transactions on Pattern Analysis and Machine Intelligence  \textbf{41}(2),  423--443 (2019). \doi{10.1109/TPAMI.2018.2798607}

\bibitem{beery2018terra}
Beery, S., Van~Horn, G., Perona, P.: Recognition in terra incognita. In: Computer Vision -- ECCV 2018. pp. 472--489. Lecture Notes in Computer Science, Springer (2018). \doi{10.1007/978-3-030-01270-0_28}

\bibitem{bible2026fujiview}
Bible, B., Hasnaeen, N., Qi, H.: {FujiView}: Multimodal late-fusion for predicting scenic visibility. In: Proceedings of the IEEE/CVF Winter Conference on Applications of Computer Vision (WACV). pp. 5653--5661 (2026). \doi{10.1109/WACV61042.2026.00548}

\bibitem{chu2017camera}
Chu, W.T., Zheng, X.Y., Ding, D.S.: Camera as weather sensor: Estimating weather information from single images. Journal of Visual Communication and Image Representation  \textbf{46},  233--249 (2017). \doi{10.1016/j.jvcir.2017.04.002}

\bibitem{glahn1972mos}
Glahn, H.R., Lowry, D.A.: The use of model output statistics ({MOS}) in objective weather forecasting. Journal of Applied Meteorology  \textbf{11}(8),  1203--1211 (1972)

\bibitem{glasner2015hot}
Glasner, D., Fua, P., Zickler, T., Zelnik-Manor, L.: Hot or not: Exploring correlations between appearance and temperature. In: Proceedings of the IEEE International Conference on Computer Vision (ICCV). pp. 3997--4005 (2015). \doi{10.1109/ICCV.2015.455}

\bibitem{graves2011visibility}
Graves, N., Newsam, S.: Using visibility cameras to estimate atmospheric light extinction. In: Proceedings of the IEEE Workshop on Applications of Computer Vision (WACV). pp. 577--584 (2011). \doi{10.1109/WACV.2011.5711556}

\bibitem{graves2014camera}
Graves, N., Newsam, S.: Camera-based visibility estimation: Incorporating multiple regions and unlabeled observations. Ecological Informatics  \textbf{23},  62--68 (2014). \doi{10.1016/j.ecoinf.2013.08.005}

\bibitem{grinsztajn2022why}
Grinsztajn, L., Oyallon, E., Varoquaux, G.: Why do tree-based models still outperform deep learning on typical tabular data? In: Advances in Neural Information Processing Systems 35 (NeurIPS 2022), Datasets and Benchmarks Track. pp. 507--520 (2022). \doi{10.52202/068431-0037}

\bibitem{gronquist2021deep}
Gr{\"o}nquist, P., Yao, C., Ben-Nun, T., Dryden, N., Dueben, P., Li, S., Hoefler, T.: Deep learning for post-processing ensemble weather forecasts. Philosophical Transactions of the Royal Society A: Mathematical, Physical and Engineering Sciences  \textbf{379}(2194),  20200092 (2021). \doi{10.1098/rsta.2020.0092}

\bibitem{hautiere2006automatic}
Hauti{\`e}re, N., Tarel, J.P., Lavenant, J., Aubert, D.: Automatic fog detection and estimation of visibility distance through use of an onboard camera. Machine Vision and Applications  \textbf{17}(1),  8--20 (2006). \doi{10.1007/s00138-005-0011-1}

\bibitem{islam2013images}
Islam, M.T., Jacobs, N., Wu, H., Souvenir, R.: Images+weather: Collection, validation, and refinement. In: IEEE CVPR Workshop on Ground Truth (2013), \url{https://mvrl.github.io/pdf/islam2013webcamweather.pdf}

\bibitem{jacobs2007consistent}
Jacobs, N., Roman, N., Pless, R.: Consistent temporal variations in many outdoor scenes. In: Proceedings of the IEEE Conference on Computer Vision and Pattern Recognition (CVPR). pp.~1--6 (2007). \doi{10.1109/CVPR.2007.383258}

\bibitem{jocher2024yolo11}
Jocher, G., Qiu, J.: Ultralytics {YOLO11}. Software, version 11.0.0, AGPL-3.0 license (2024), \url{https://github.com/ultralytics/ultralytics}

\bibitem{kapoor2023leakage}
Kapoor, S., Narayanan, A.: Leakage and the reproducibility crisis in machine-learning-based science. Patterns  \textbf{4}(9),  100804 (2023). \doi{10.1016/j.patter.2023.100804}

\bibitem{ke2017lightgbm}
Ke, G., Meng, Q., Finley, T., Wang, T., Chen, W., Ma, W., Ye, Q., Liu, T.Y.: {LightGBM}: A highly efficient gradient boosting decision tree. In: Advances in Neural Information Processing Systems. vol.~30. Curran Associates, Inc. (2017), \url{https://proceedings.neurips.cc/paper_files/paper/2017/file/6449f44a102fde848669bdd9eb6b76fa-Paper.pdf}

\bibitem{kim2022shortterm}
Kim, B.Y., Belorid, M., Cha, J.W.: Short-term visibility prediction using tree-based machine learning algorithms and numerical weather prediction data. Weather and Forecasting  \textbf{37}(12),  2263--2274 (2022). \doi{10.1175/WAF-D-22-0053.1}

\bibitem{koh2021wilds}
Koh, P.W., Sagawa, S., Marklund, H., Xie, S.M., Zhang, M., Balsubramani, A., Hu, W., Yasunaga, M., Phillips, R.L., Gao, I., Lee, T., David, E., Stavness, I., Guo, W., Earnshaw, B., Haque, I., Beery, S.M., Leskovec, J., Kundaje, A., Pierson, E., Levine, S., Finn, C., Liang, P.: {WILDS}: A benchmark of in-the-wild distribution shifts. In: Proceedings of the 38th International Conference on Machine Learning. Proceedings of Machine Learning Research, vol.~139, pp. 5637--5664. PMLR (2021), \url{https://proceedings.mlr.press/v139/koh21a.html}

\bibitem{kopecka2026estimation}
Kopeck{\'a}, J., Kopeck{\'y}, D., {\v{S}}tursa, D., R{\'a}cov{\'a}, Z., Krej{\v{c}}{\'i}, T., Dole{\v{z}}el, P.: Estimation of atmospheric visibility by deep learning model using multimodal dataset. Knowledge-Based Systems  \textbf{331},  114732 (2026). \doi{10.1016/j.knosys.2025.114732}

\bibitem{kurashige1998howmany}
Kurashige, Y., Miyashita, A.: How many days in a year can {Mt. Fuji} and the {Tokyo Tower} be seen from the {Tokyo} suburban area? Journal of the Air \& Waste Management Association  \textbf{48}(8),  763--765 (1998). \doi{10.1080/10473289.1998.10463724}

\bibitem{laffont2014transient}
Laffont, P.Y., Ren, Z., Tao, X., Qian, C., Hays, J.: Transient attributes for high-level understanding and editing of outdoor scenes. ACM Transactions on Graphics  \textbf{33}(4),  1--11 (2014). \doi{10.1145/2601097.2601101}

\bibitem{lepetit2020learning}
Lepetit, P., Barth{\`e}s, L., Mallet, C., Viltard, N.: Learning to compare visibility on webcam images. In: Proceedings of the 10th International Conference on Climate Informatics (CI2020). pp. 91--97. ACM (2020). \doi{10.1145/3429309.3429323}

\bibitem{li2017meteorological}
Li, S., Fu, H., Lo, W.L.: Meteorological visibility evaluation on webcam weather image using deep learning features. International Journal of Computer Theory and Engineering  \textbf{9}(6),  455--461 (2017). \doi{10.7763/IJCTE.2017.V9.1186}

\bibitem{liang2024foundations}
Liang, P.P., Zadeh, A., Morency, L.P.: Foundations \& trends in multimodal machine learning: Principles, challenges, and open questions. ACM Computing Surveys  \textbf{56}(10),  1--42 (2024). \doi{10.1145/3656580}

\bibitem{lo2025meteorological}
Lo, W.L., Wong, K.W., Hsung, R.T.C., Chung, H.S.H., Fu, H.: Meteorological visibility estimation using landmark object extraction and the {ANN} method. Sensors  \textbf{25}(3), ~951 (2025). \doi{10.3390/s25030951}

\bibitem{ma2021smil}
Ma, M., Ren, J., Zhao, L., Tulyakov, S., Wu, C., Peng, X.: {SMIL}: Multimodal learning with severely missing modality. In: Proceedings of the AAAI Conference on Artificial Intelligence. vol.~35, pp. 2302--2310 (2021). \doi{10.1609/aaai.v35i3.16330}

\bibitem{meyer2018improving}
Meyer, H., Reudenbach, C., Hengl, T., Katurji, M., Nauss, T.: Improving performance of spatio-temporal machine learning models using forward feature selection and target-oriented validation. Environmental Modelling \& Software  \textbf{101}, ~1--9 (2018). \doi{10.1016/j.envsoft.2017.12.001}

\bibitem{murphy1993good}
Murphy, A.H.: What is a good forecast? {An} essay on the nature of goodness in weather forecasting. Weather and Forecasting  \textbf{8}(2),  281--293 (1993)

\bibitem{neverova2016moddrop}
Neverova, N., Wolf, C., Taylor, G., Nebout, F.: {ModDrop}: Adaptive multi-modal gesture recognition. IEEE Transactions on Pattern Analysis and Machine Intelligence  \textbf{38}(8),  1692--1706 (2016). \doi{10.1109/TPAMI.2015.2461544}

\bibitem{norouzzadeh2018automatically}
Norouzzadeh, M.S., Nguyen, A., Kosmala, M., Swanson, A., Palmer, M.S., Packer, C., Clune, J.: Automatically identifying, counting, and describing wild animals in camera-trap images with deep learning. Proceedings of the National Academy of Sciences  \textbf{115}(25),  E5716--E5725 (2018). \doi{10.1073/pnas.1719367115}

\bibitem{palvanov2019visnet}
Palvanov, A., Cho, Y.I.: {VisNet}: Deep convolutional neural networks for forecasting atmospheric visibility. Sensors  \textbf{19}(6), ~1343 (2019). \doi{10.3390/s19061343}

\bibitem{rasp2018neural}
Rasp, S., Lerch, S.: Neural networks for postprocessing ensemble weather forecasts. Monthly Weather Review  \textbf{146}(11),  3885--3900 (2018). \doi{10.1175/MWR-D-18-0187.1}

\bibitem{reichstein2019deep}
Reichstein, M., Camps-Valls, G., Stevens, B., Jung, M., Denzler, J., Carvalhais, N., Prabhat: Deep learning and process understanding for data-driven {Earth} system science. Nature  \textbf{566}(7743),  195--204 (2019). \doi{10.1038/s41586-019-0912-1}

\bibitem{richardson2018tracking}
Richardson, A.D., Hufkens, K., Milliman, T., Aubrecht, D.M., Chen, M., Gray, J.M., Johnston, M.R., Keenan, T.F., Klosterman, S.T., Kosmala, M., Melaas, E.K., Friedl, M.A., Frolking, S.: Tracking vegetation phenology across diverse {North American} biomes using {PhenoCam} imagery. Scientific Data  \textbf{5},  180028 (2018). \doi{10.1038/sdata.2018.28}

\bibitem{roberts2017crossvalidation}
Roberts, D.R., Bahn, V., Ciuti, S., Boyce, M.S., Elith, J., Guillera-Arroita, G., Hauenstein, S., Lahoz-Monfort, J.J., Schr{\"o}der, B., Thuiller, W., Warton, D.I., Wintle, B.A., Hartig, F., Dormann, C.F.: Cross-validation strategies for data with temporal, spatial, hierarchical, or phylogenetic structure. Ecography  \textbf{40}(8),  913--929 (2017). \doi{10.1111/ecog.02881}

\bibitem{seyednasrollah2019phenocam}
Seyednasrollah, B., Young, A.M., Hufkens, K., Milliman, T., Friedl, M.A., Frolking, S., Richardson, A.D.: Tracking vegetation phenology across diverse biomes using version 2.0 of the {PhenoCam} dataset. Scientific Data  \textbf{6}, ~222 (2019). \doi{10.1038/s41597-019-0229-9}

\bibitem{shwartzziv2022tabular}
Shwartz-Ziv, R., Armon, A.: Tabular data: Deep learning is not all you need. Information Fusion  \textbf{81},  84--90 (2022). \doi{10.1016/j.inffus.2021.11.011}

\bibitem{snoek2005early}
Snoek, C.G.M., Worring, M., Smeulders, A.W.M.: Early versus late fusion in semantic video analysis. In: Proceedings of the 13th Annual ACM International Conference on Multimedia (MM '05). pp. 399--402. ACM (2005). \doi{10.1145/1101149.1101236}

\bibitem{strobl2007bias}
Strobl, C., Boulesteix, A.L., Zeileis, A., Hothorn, T.: Bias in random forest variable importance measures: Illustrations, sources and a solution. BMC Bioinformatics  \textbf{8}(1), ~25 (2007). \doi{10.1186/1471-2105-8-25}

\bibitem{swanson2015snapshot}
Swanson, A., Kosmala, M., Lintott, C., Simpson, R., Smith, A., Packer, C.: Snapshot {Serengeti}, high-frequency annotated camera trap images of 40 mammalian species in an {African} savanna. Scientific Data  \textbf{2},  150026 (2015). \doi{10.1038/sdata.2015.26}

\bibitem{wang2020multimodal}
Wang, H., Shen, K., Yu, P., Shi, Q., Ko, H.: Multimodal deep fusion network for visibility assessment with a small training dataset. IEEE Access  \textbf{8},  217057--217067 (2020). \doi{10.1109/ACCESS.2020.3031283}

\bibitem{wang2020what}
Wang, W., Tran, D., Feiszli, M.: What makes training multi-modal classification networks hard? In: Proceedings of the IEEE/CVF Conference on Computer Vision and Pattern Recognition (CVPR). pp. 12692--12702 (2020). \doi{10.1109/CVPR42600.2020.01271}

\bibitem{yao2022wildtime}
Yao, H., Choi, C., Cao, B., Lee, Y., Koh, P.W., Finn, C.: Wild-time: A benchmark of in-the-wild distribution shift over time. In: Advances in Neural Information Processing Systems 35 (NeurIPS 2022), Datasets and Benchmarks Track. pp. 10309--10324 (2022). \doi{10.52202/068431-0749}

\bibitem{you2019relative}
You, Y., Lu, C., Wang, W., Tang, C.K.: Relative {CNN}-{RNN}: Learning relative atmospheric visibility from images. IEEE Transactions on Image Processing  \textbf{28}(1),  45--55 (2019). \doi{10.1109/TIP.2018.2857219}

\bibitem{zippenfenig2024openmeteo}
Zippenfenig, P.: Open-meteo.com weather api. Software, version 1.4.0 (2024). \doi{10.5281/zenodo.7970649}, \url{https://open-meteo.com}

\end{thebibliography}

\end{document}